Disclaimer: This preprint has now been peer-reviewed and published in the following journal. Please refer to the final published version for citation purpose:

Lee, Y. K., Lee, I., Shin, M., Bae, S., & Hahn, S. (2024). Enhancing Empathic Reasoning of Large Language Models Based on Psychotherapy Models for AI-assisted Social Support. Korean Journal of Cognitive Science, 35(1). http://dx.doi.org/10.19066/cogsci.2024.35.1.002

# Chain of Empathy: Enhancing Empathetic Response of Large Language Models Based on Psychotherapy Models

**Yoon Kyung Lee, Inju Lee, Minjung Shin, Seoyeon Bae, & Sowon Hahn**

Human Factors Psychology Lab
Seoul National University
yoonlee78@snu.ac.kr, swhahn@snu.ac.kr

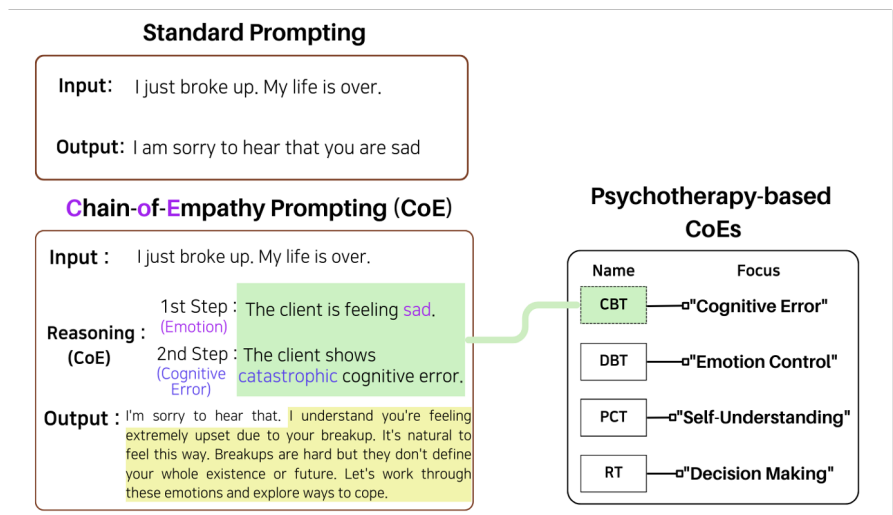

Figure 1: Chain-of-Empathy (CoE) prompting with cognitive reasoning of human's emotion based on psychotherapy models.

## Abstract

We present a novel method, the Chain of Empathy (CoE) prompting, that utilizes insights from psychotherapy to induce Large Language Models (LLMs) to reason about human emotional states. This method is inspired by various psychotherapy approaches—Cognitive-Behavioral Therapy (CBT), Dialectical Behavior Therapy (DBT), Person-Centered Therapy (PCT), and Reality Therapy (RT)—each leading to different patterns of interpreting clients' mental states. LLMs without reasoning generated predominantly exploratory responses. However, when LLMs used CoE reasoning, we found a more comprehensive range of empathetic responses aligned with each psychotherapy model's different reasoning patterns. The CBT-based CoE resulted in the most balanced generation of empathetic responses. The findings underscore the importance of understanding the emotional context and how it affects human-AI communication. Our research contributes to understanding how psychotherapeutic models can be incorporated into LLMs, facilitating the development of context-specific, safer, and empathetic AI.

## 1. Introduction

Large Language Models (LLMs) have dramatically improved text generation performance that highly resembles human expressions (Brown et al., 2020; Touvron et al., 2023; Taori et al., 2023; Bommasani et al., 2021). These models have been showcasing their reasoning abilities and achieving high performance in various problem-solving tasks, including professional exams such as the bar exam (Bommarito II and Katz, 2022), a math test (Zhang et al., 2023), and medical diagnoses (Nori et al., 2023). Among many recent findings related to LLMs, one interesting point is the introduction of 'Chain-of-Thought (CoT)' prompting (Wei et al., 2022; Kojima et al., 2022). This method elicits reasoning before generating outputs. Nevertheless, this recent method has primarily experimented with logical or arithmetic tasks. Whether reasoning about emotional states or underlying causes enhances empathetic responses to user input remains a relatively under-explored area and merits investigation.

Empathetic understanding requires cognitive reasoning of others' mental states. Different psychotherapeutic approaches offer varied perspectives on empathy (Hofmann et al., 2010; Linehan, 1987; Cooper and McLeod, 2011; Wubbolding et al., 2017). By integrating these approaches into LLMs' reasoning stage, we can enhance the depth and specificity of their empathetic responses. For this purpose, this study delves into these possibilities and proposes a novel prompting, Chain-of-Empathy prompting (CoE). The CoE prompt integrates a reasoning process into text generation. It focuses on clients' emotions and the specific factors leading to those emotions, such as cognitive errors, before generating the output.

## 2. Related Work

### 2.1. Theoretical Backgrounds of Empathy

Empathy, defined as sharing others' emotions and experiences, is a multifaceted concept encompassing cognitive and emotional aspects (Neff, 2003; Anderson and Keltner, 2002; De Vignemont and Singer, 2006; Hall and Schwartz, 2019; Zaki, 2019). Cognitive empathy involves understanding others' emotions and perspectives, linked to abilities such as mentalizing and narrative imagination (Eisenberg, 2014). It requires an in-depth cognitive appraisal of the situation, considering factors like pleasantness, control, and certainty of the outcome (Lazarus, 1991; Wondra and Ellsworth, 2015). Affective (emotional) empathy allows individuals to experience others' emotions, while motivational empathy, a newer concept, embodies the desire to alleviate others' emotional distress (Zaki, 2019).

### 2.2. Empathetic Communication in Text

Natural Language Processing (NLP) has been increasingly employed in developing conversational agents, or chatbots, across various professional domains. These include mental healthcare for victims of crime (Ahn et al., 2020), individuals on the autism spectrum (Diehl et al., 2012), and those suffering from anxiety disorders (Rasouli et al., 2022).

Recently, chatbots designed for psychotherapy (e.g., CBT) have shown promising results in assisting the long-term treatment of anxiety and depression (Nwosu et al., 2022). However, current AI-generated responses appear generic and less authentic, making personalized responses a significant challenge. Empathetic reasoning is crucial for these systems, leading to ongoing efforts to enhance their empathetic expression capabilities by incorporating human-like traits (Roller et al., 2021).

### 2.3. Computational Approach to Empathy

Past research in psychotherapy has primarily focused on empathy based on the analysis of nonverbal cues, such as body language and facial expressions, often requiring manual coding of empathetic responses (Scherer et al., 2001; American Psychiatric Association et al., 1994; Ekman and Friesen, 1971).

Recent advances in artificial intelligence have shifted towards a computational approach, where empathy is predicted from a text corpus and quantified through the labeling of emotions (Rashkin et al., 2019) and distress (Buechel et al., 2018). While most studies have traditionally concentrated on the client's capacity for empathy, the expressed empathy of the counselor is increasingly recognized as critical to successful therapy outcomes (Truax and Carkhuff, 2007). This aspect of expressed empathy is particularly relevant to our approach, where we aim to use LLMs to reflect their understanding of the client's needs accurately.

### 2.4. Reasoning in Large Language Models

Recently, CoT has shown effective in eliciting the reasoning process of the LLMs (Wei et al., 2022; Prystawski et al., 2022; Yao et al., 2023; Kojima et al., 2022). CoT prompting in previous research has included reasoning steps within the prompt instruction for zero- or one-shot learning of LLMs during text generation (Kojima et al., 2022). This model has improved the performance of problem-solving (Kojima et al., 2022) or metaphor understanding (Prystawski et al., 2022), offering new insights and suggesting possibilities for generative models to be used in many other domains.

| | Prompt conditions | | | |
|---|---|---|---|---|
| | **CBT-CoE** | **DBT-CoE** | **PCT-CoE** | **RT-CoE** |
| **Goal** | Cognitive reframing | Emotion regulation | Self-understanding | Problem-focused coping |
| **Reasoning** | Tackling negative thought patterns | Addressing emotional dysregulation | Enhancing self-awareness | Identifying cause of the dissatisfaction |

Table 1: Comparison of goals and reasoning style in different psychotherapy based CoEs.

## 3. The Present Study

We investigated whether eliciting empathetic reasoning in LLMs leads to natural responses. Therefore, we developed CoE prompting to reason emotion and situational factors that could help the model to accurately infer the client's emotional experience in mental healthcare and thus choose the most appropriate and context-aware empathetic strategy to communicate.

## 4. Methods

### 4.1. Language Model

We used GPT-3.5 API from OpenAI [1] for system setup. The model ('text-davinci-003') temperature was set to 0.9. The top p parameter was set to 1 for nucleus sampling to reduce the randomness of the output (The frequency penalty = 0 and the presence penalty = 0.6).

### 4.2. Chain-of-Empathy Reasoning

Table 1 and Figure 1 show four unique prompts with CoE in addition to the base condition (no reasoning): Cognitive-Behavioral Therapy (CBT; Beck, 1979; Kaczkurkin and Foa, 2022; Hofmann et al., 2010), Dialectical Behavior Therapy (DBT; Linehan, 1987), Person-Centered Therapy (PCT; Cooper and McLeod, 2011; Knutson and Koch, 2022), and Reality Therapy (RT; Wubbolding et al., 2017)[2]. Except for the base condition, these prompts' instructions were designed to reflect the therapists' reasoning process in their respective counseling models.

Models in each prompting condition were tested in zero-shot, with only instructions on which option to choose per class: empathetic strategy (emotional reaction, exploration, and interpretation) and communication level (no expression, weak, and strong) (Sharma et al., 2020). The common reasoning steps involved in each CoE condition were: (1) Identify any word that represents the client's emotion, and (2) Understand the individual/situational factors that may have led to the expression in the client's message.

## 5. Experiments

We instructed each LLM to generate appropriate responses to the posts of seekers seeking advice on Reddit and predict the best suitable empathetic strategy. For the ground-truth label of each empathetic strategy class, we used the EPITOME [3], crowdsourced Reddit posts of mental health, with an average inter-annotator agreement reported as above 0.68 (Sharma et al., 2020). The dataset comprised pairs of help-seeking posts and responding posts. Each pair was labeled based on (1) the type of expressed "empathy mechanism" (i.e., empathy strategy) and (2) the presence and "level" of each expressed empathy (i.e., communication strength). The three empathy strategies included emotional reaction, exploration, and interpretation, with corresponding levels of 0, 1, and 2. Pairs labeled as level 0, indicating no expression of empathy, were excluded. The number of pairs for each strategy was as follows: 'emotion reaction'=1,047, 'exploration'=481, and 'interpretation'=1,436. We randomly sampled 500 pairs in each emotional reaction and interpretation data to balance the number of pairs between strategies. Each strategy's final number of pairs was emotional reaction=500, exploration=480, and interpretation=500.

---

[1] https://openai.com/
[2] We want to emphasize that these descriptions are not exhaustive representations of the goals of each psychotherapy. These goals and reasoning strategies have been specifically modified for LLM prompting and do not reflect the entire interaction between clinical/counseling psychologists and clients.
[3] https://github.com/behavioral-data/ Empathy-Mental-Health

|  | Acc | Emotional Reaction | | | Interpretation | | | Exploration | | |
| --- | --- | --- | --- | --- | --- | --- | --- | --- | --- | --- |
|  |  | Prec. | Recall | F1 | Prec. | Recall | F1 | Prec. | Recall | F1 |
| **Base** | 0.340 | 0.467 | 0.185 | 0.27 | 0 | 0 | 0 | 0.327 | 0.866 | 0.475 |
| **CBT-CoE** | 0.319 | 0.463 | 0.165 | 0.244 | 0.293 | 0.260 | 0.276 | 0.303 | 0.543 | 0.389 |
| **DBT-CoE** | 0.334 | 0.392 | 0.372 | 0.382 | 0.291 | 0.060 | 0.100 | 0.309 | 0.582 | 0.404 |
| **PCT-CoE** | 0.336 | 0.399 | 0.243 | 0.302 | 0.333 | 0.016 | 0.031 | 0.319 | 0.757 | 0.449 |
| **RT-CoE** | 0.336 | 0.407 | 0.308 | 0.350 | 0.354 | 0.044 | 0.079 | 0.309 | 0.664 | 0.420 |

Table 2: Model performance in empathetic strategy classification task by CoE prompting conditions. *Prec. = Precision

## 5.1. Model Performances

Table 2 and Figure 2 show the performance of the empathetic strategy classification of LLMs with each CoE prompt, measured in terms of precision, recall, F1 score, and accuracy. Upon generating a response, each model with CoE prompts predicted which empathy strategy is most suitable for each seeker's post among the three strategies. We then compared the predicted empathy strategy with the ground truth empathy strategy and calculated prediction accuracy.

Outputs with errors in the predicted strategy names were excluded from the analysis. Most of these errors resulted from the nature of LLM as a generative model, which behaves differently from traditional supervised learning models for classification tasks. Despite explicit instructions in the prompt, the models occasionally generated 'noise' output and predicted strategies that were not among the provided options. These errors include responses of failed predictions or response retrieval (e.g., "No Empathy Strategy"). In addition, they sometimes predicted new strategies which did not fall into any of the predefined three strategies (e.g., "Reflection," "Validation: acknowledging the client's feelings and experiences," and "Approval: expressing approval or positive reinforcement to the client").

## 6. Qualitative Evaluations

The LLM generally generated courteous and comprehensive responses. While many human peer supporters often provided brief comments and shared personal opinions or give advice, the CoE LLM mostly responded with at least two empathetic strategies and frequently suggested seeking professional help. The model tended to initiate responses by interpreting users' current state and subsequent advice or exploring potential options. For example, when a distressed seeker could not control her anxiety after a violent fight between her parents, DBT-CoE prompt responded with multiple empathetic strategies, "I'm so sorry you had to witness that. It's understandable that you're feeling overwhelmed and scared right now. It's not okay for anyone to threaten or hurt another person, and it's not your fault. How can I support you right now?". This contradicts the original human response in benchmark data: "Everything is wrong with people."

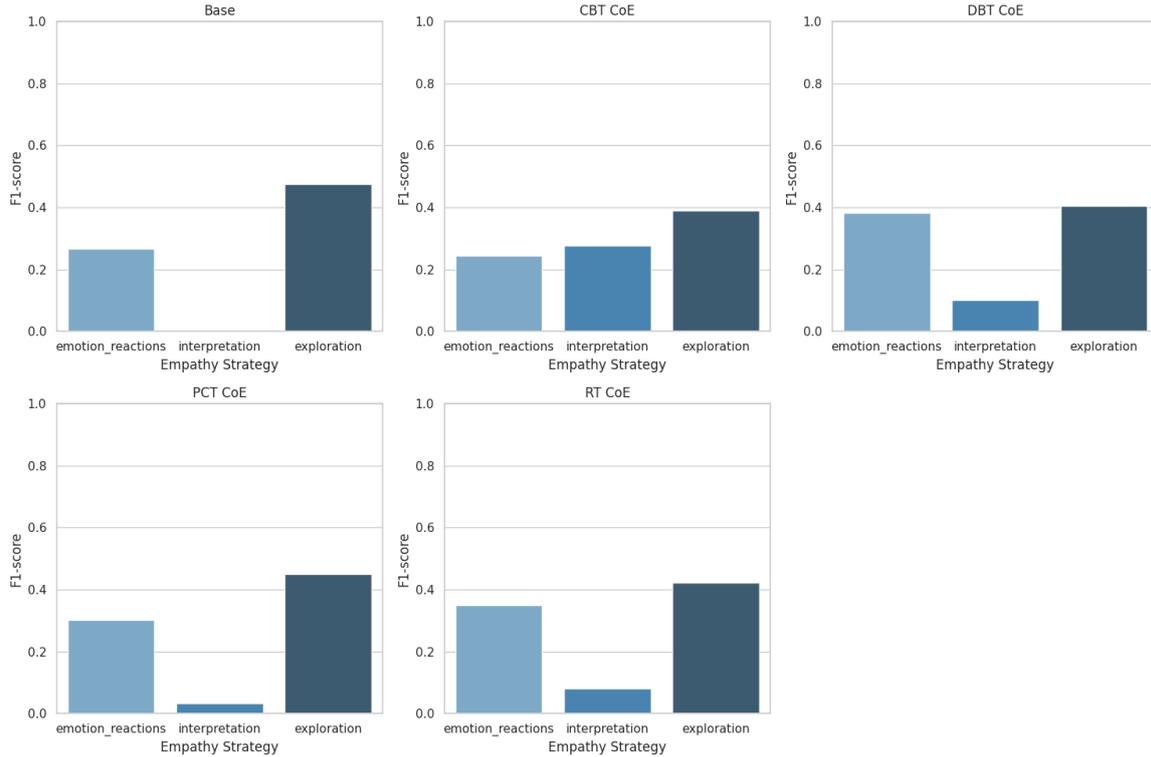

Figure 2: Empathic expression strategy classification accuracy per prompt conditions. Compared to Base condition, CBT-CoE provided the balanced set of each empathy expression but less emotional reaction than other CoEs.

## 7. Conclusions

In summary, we developed a CoE reasoning prompt for generating empathetic responses based on psychotherapy models, and we compared the performance of empathetic strategy classification. Our findings revealed that LLMs without reasoning showed a significant preference for the exploration strategy, with interpretation being the least preferred strategy. Although all reasoning prompts generated responses most strongly associated with exploration, they differed from the base prompt by generating interpretation to a certain extent. Intriguingly, only the CBT-CoE generated the highest number of the interpretation strategy. This pattern might reflect CBT's inherent approach - clarifying cognitive errors to clients. These findings highlight the importance of incorporating context-specific empathetic reasoning in therapeutic interactions with generative AIs.

## 8. Limitations and Suggestions

We acknowledge several limitations that should be considered in future research and development. First, we did not employ more extensive evaluative criteria for empathy, especially those validated from psychology literature like *the Interpersonal Reactivity Index* (Davis, 1980; Davis, 1983). Future studies should consider evaluating LLMs using these established scales to assess their communication skills for validity and reproducibility.

Our evaluation focused solely on the empathic accuracy of the LLMs' and did not measure user perception. User perception of empathetic expression varies depending on whether they interact with humans or artificially intelligent systems (Medeiros et al., 2021). Furthermore, people perceive and react differently to AIs' empathetic expressions (Urakami et al., 2019). Thus, future works should investigate how users perceive and respond to the models' empathetic responses to enhance our understanding of the efficacy of LLMs' empathetic expressions.

For quantitative evaluation, we used a single LLM model (GPT-3.5) and one domain, mental health. Incorporating a diverse text corpus, including career coaching and motivational interviewing (Miller and Rollnick, 2012), could enable LLMs to produce more personalized communication. This presents an opportunity for future research to encompass a wider array of topics and conversational styles, thereby increasing the reliability of LLM's performance. Additionally, different LLMs may excel in varied capabilities, leading each LLM to display optimal performance in specific tasks (Sivarajkumar et al., 2023). Investigating and assessing the empathetic expressions generated by different LLMs is crucial for a comprehensive evaluation of LLMs' ability to discern human emotions and craft appropriate, empathetic responses.

## 9. Ethical Considerations

The expanding use of large language models (LLMs), especially within mental healthcare, calls for thoughtful ethical engagement. As these models advance in generating responses that mirror human counselors, it is imperative we closely examine their impact on users, particularly those navigating mental health challenges.